\documentclass{ceurart}
\usepackage{graphicx}
\begin{document}

\copyrightyear{2023}
\copyrightclause{Copyright for this paper by its authors.
  Use permitted under Creative Commons License Attribution 4.0
  International (CC BY 4.0).}

\conference{De-Factify 3.0: Workshop on Multimodal Fact Checking and Hate Speech Detection, co-located with AAAI 2024}

\title{Ontology Enhanced Claim Detection}

%\author[1]{Author 1}[
%info
%]
%\author[2]{Author 2}[
%info
%]
%\address[1]{address info}
%\address[2]{address info}

\author[1]{Zehra Melce Hüsünbeyi}[%
email=melce.husunbeyi@rub.de
]
\author[1]{Tatjana Scheffler}[%
email=tatjana.scheffler@rub.de,
url=https://staff.germanistik.rub.de/digitale-forensische-linguistik/
]
\address[1]{Digital Forensic Linguistics, Ruhr-Universität-Bochum,
  Universitätsstraße 150, 44801 Bochum, Germany}
\begin{abstract}
We propose an ontology enhanced model for sentence based claim detection. We fused ontology embeddings from a knowledge base with BERT sentence embeddings to perform claim detection for the ClaimBuster and the NewsClaims datasets. Our ontology enhanced approach showed the best results with these small-sized unbalanced datasets, compared to  other statistical and neural machine learning models. The experiments demonstrate that adding domain specific features (either trained word embeddings or knowledge graph metadata) can improve traditional ML methods. In addition, adding domain knowledge in the form of ontology embeddings helps avoid the bias encountered in neural network based models, for example the pure BERT model bias towards larger classes in our small corpus. 
\end{abstract}


\maketitle

\section{Introduction}
With the enormous increase in internet resources, the verification of information on the internet has gained importance. On the one hand,  controlling the dissemination of (false) information manually is not sustainable, because it requires too much human work at large scales. On the other hand, fully automatic methods based on text alone lack domain knowledge and may introduce erroneous decisions.  In this context, knowledge-based approaches have been studied, particularly  on data mining, verification, and representation of factual knowledge from the web. %Within this approach, the task of claim identification, the first stage to automating disinformation detection and assist fact-checking organizations, can be examined based on both content and context. 
We utilize this approach to address claim identification as a sentence classification problem, the first step in order to automate disinformation detection and assist fact checking organizations, based on both content and the larger context. For this purpose, we fused the modalities of textual information and knowledge based representations.

Achieving a common understanding and terminology has become an essential difficulty in the claim identification process. Identifying and defining the fact-check worthiness of a claim is influenced by the subjectivity of the problem, the background of the annotator, the context of expression in political discourse, and the inherent bias of the fact-checking organization \citep{boland2022beyond}. In the past, the notion of a relevant ``claim'' has been operationalized in different ways, for example, whether a statement is in the public interest, is a rumor or not, or is verifiable or not. In this paper, we studied on the ClaimBuster \citep{arslan2020benchmark} and NewsClaims \citep{gangi-reddy-etal-2022-newsclaims} datasets. ClaimBuster data uses the concept of check-worthiness and in the NewsClaims dataset, researchers extend this approach with including claim related attributes, (i.e., object of the claim). %This dataset captures more general issues and patterns associated with assessing check-worthy factual claims by covering presidential election speeches over 50 years. 
%% TS: I removed this sentence because the introduction is not the right place for this. 
%We thus operationalize the task as a sentence classification problem as checkworthy claim, non-checkworthy claim, or non-factual statement.

%a difficult task to conceptualize and requires both expression and features such as place, time, and agent, 
A specific challenge for machine learning approaches to fact-checking and claim identification is the lack of realistic data sets for many languages. While disinformation is an international problem relevant to a range of languages, high quality, large-scale data sets for languages other than English are almost nonexistent. When they exist, they are very small, leading to gaps in available information and poor results. In addition to pure textual features, we therefore propose to combine  information obtained from knowledge bases and ontologies into the model.
%In order to capture such a difficult conceptualization in machine-readable format, we used
%In order to address this difficult problem, w
We compared statistical machine learning and neural network based methods, and additionally integrated structured web resources and linguistic features to assess their value in this domain. The meta-features presented in the ClaimBuster dataset were embedded using the Translating Embeddings for Modeling Multi-relational Data (TransE) approach \citep{bordes2013translating}. In addition, ClaimsKG data \citep{tchechmedjiev2019claimskg}, which is a structured database for claims, was scraped and an OWL ontology \citep{antoniou2009web} was created to be used in the claim detection task. The embedded OWL ontology containing the ClaimReview data of the web\footnote{\url{https://toolbox.google.com/factcheck/apis}} was used to encode the ClaimBuster and NewsClaims datasets. These were fine-tuned with the BERT model \citep{devlin2018bert} as a contextual embedding method. We show that an ontology enhanced conceptual model can be successful in addressing a challenging task that  includes implicit meanings. In our experiments,  models incorporating some domain information (notably, ontological resources) perform better than pure NLP models. 

%In the following sections, we describe 

%"Depending on the use-case and the model, any one of them might be able to do the required task"
Our contributions can be summarized as follows: We create an OWL ontology based on the \mbox{ClaimsKG} dataset of claims from fact-checking organizations. We systematically explore the performance of statistical and neural machine learning models for claim identification, using a variety of lexical, linguistic, and ontological features. We propose an improved approach for fusing domain-specific features such as the OWL ontology into neural methods. Finally, we demonstrate the effectiveness of ontology information in claim identification in small datasets.

%-creation of OWL ontology by taking advantage of claimsKG data\\ 
%-proposing improved approach by fusing domain-specific features into neural models\\
%-examining the effect of several vectorization methods along with statistical ML methods and comparison with neural network-based approach on a small domain-specific dataset 

\section{Related Work} \label{related}
Recent research on the claim identification task can be mainly categorized as feature engineering based approaches and neural network based ones. In the literature, statistical machine learning classifiers are commonly combined with surface level features such as named entities, verbal forms, bag-of-words feature, and part-of-speech (POS) tags. \citet{zuo2018hybrid} offered lexical, stylometric and semantic features for political debates and speeches in a hybrid approach by combining rules and supervised learning approaches and they achieved promising results. The most discriminative features presented by \citet{hassan} were POS tags and entity types along with an SVM classifier. Another approach that combined sentence embeddings (i.e., InferSent embeddings) and POS tags with named entities in a Logistic Regression classifier  achieved better results with universal sentence representations than the word-level representations \citep{konstantinovskiy2021toward}. 

%In the literature, statistical machine learning classifiers are commonly combined with surface level features such as lexical and syntactic features (Zhou et al., 2020), named entities and verbal forms for political transcripts (Zuo et al., 2018), or bag-of-words features, part-of-speech tags and entity types (Arslan, et al., 2017). There is also an approach that combines sentence embeddings (i.e., InferSent embeddings) and POS tags with named entities in a Logistic Regression classifier (Konstantinovskiy et al., 2021). \textcolor{red}{RESULTS OR CONCLUSION??} 
%TS: What do these models show? What is your conclusion or the authors' conclusion from these approaches? eg: These models show that lexical features are useful... that syntactic properties can improve... etc. 
% How did these models inform your work??
% ALSO: numerical results? On xxx corpus they achieved a yyy score for two-way classification (claim/non-claim)... give some details on how large the data was, type of data, etc.

Neural network-based models have also been used to address the challenge of automatically detecting claims. \citet{jimenez2018empirical} provided an enhanced version of the ClaimBuster dataset by adding random sequences of text that included mostly numbers to increase robustness of the model and learn from negative instances. They proposed an architecture consisting of convolutional and pooling layers followed by a Bi-directional LSTM network by taking advantage of pre-trained word embeddings. Experiment results showed different combinations of word embeddings did not improve the models significantly and the addition of nonsensical data has no negative impact on the regular dataset. In another approach, a neural check-worthiness sentence ranking model was offered by performing two distinct word encodings such as one hot encoded syntactic dependencies of words and trained word embeddings using the Word2Vec skip-gram model \citep{hansen2019neural}. These input representations were fed to a RNN based neural network model with gated recurrent units and aggregated outputs followed by an attention mechanism. The researchers stated that they made the first contribution to examining the effect of word embeddings trained on a domain-specific dataset for this task. Their performance scores achieved large improvements with domain-specific embeddings and syntactic dependency parsing. \citep{meng2020gradient} performed BERT and gradient-based adversarial training on the binary version of ClaimBuster dataset. To the best of our knowledge, the most recent work on the ClaimBuster dataset including both crowdsourced and groundtruth parts by considering checkworthiness and the three class distribution has been conducted by \citet{jha2023towards}. They achieved best performance with Glove embedding and RNN based approach. 

%ClaimBuster studies... TODO
%\citep{jha2023towards}, \citep{gangi-reddy-etal-2022-newsclaims}, \citep{meng2020gradient}**, 
% you do not need to mention all or too many!! just a couple is ok

Existing models were applied to the claim detection problem with different conceptualizations and different datasets. Even though it is difficult to establish what the state-of-the-art model for the claim detection task is, we can conclude that domain-specific feature enhancement is a significant approach for implementing promising models. As stated in the survey study \citep{sikelis2021ontology}, ontologies provide an accurate representation of the problem domain with a vast amount of information from the web and a rich set of axioms to link pieces of information. To this end, \citet{tchechmedjiev2019claimskg} published a knowledge graph of fact-checked claims, ClaimsKG, including English claims and related metadata such as their labels, authors, dates, debunked news, keywords extracted from the claim body and from the review body together with their Wikipedia categories, and references cited in the claim reviews. We aim to take advantage of this graph-structured data to create a claim ontology and obtain its vector representation  to enhance the neural network-based model. To the best of our knowledge, ours is the first study combining ontology embeddings and BERT to identify check-worthy claims.

\section{Data and task}
%\subsection{Dataset}
\paragraph{ClaimBuster} is a publicly available dataset created to address the claim identification task, which is an understudied problem, addressing the concept of checkworthiness \citep{arslan2020benchmark}. This corpus includes sentence-level annotations of claims from U.S. presidential debates, from 1960 to 2016. Each sentence has been marked with one of three categories: non-factual statement (NFS), unimportant factual statement (UFS), and check-worthy factual statement (CFS). The sentences belonging to the CFS class include facts that the public would like to know about their validity. Sentences containing facts where the public doesn't question their validity belong to the UFS class. The NFS class consists of sentences that do not contain any factual assertions, such as subjective sentences and interrogative sentences.

We used both the ground-truth corpus containing 1032 sentences annotated by three experts, and the crowdsourced corpus with 22,501 sentences that were labeled by 2 or 3 participants. For small data experiments, one-third of the ground-truth dataset was allocated as the test set throughout all experiments. Thus, the effect of feature sets in small and content-dependent datasets on neural network-based and statistical ML approaches can be examined using this corpus. %We chose only the smaller, expert annotated subset of the corpus to ensure data quality and because future work on different languages will in any case have to cope with small datasets. 
As continuation of the study, the models were trained on the crowdsourced part of the dataset and tested with the groundtruth part. 
\paragraph{NewsClaims} was published as a new evaluation benchmark dataset for the claim detection task. This dataset consists of news articles related to COVID-19 in the LDC corpus LDC2021E11. 889 claims from 143 news articles were annotated by 3 annotators in terms of claim validity as well as other attributes such as claim topic, claim object, span, claimer, and stance. As stated by the researchers, NewsClaims is suitable for the claim identification task in a low resource setting due to the lack of previous data and urgency of the situation.

For claim identification, we extracted sentences labeled as factually verifiable and considered the remaining sentences of the articles as counter instances. To decrease the noise of the dataset, duplications, URLs, citations and sentences including less than 4 tokens were excluded. We used one third of the 5157 binary labeled sentences as test data in our proposed models for the sake of consistency.

Table~\ref{distribution} shows the class imbalance in the corpora as is observed in almost all real-life datasets. 
% TO DO: add class based distribution info for all datasets (claimBuster-small/large and newsClaims)
% \usepackage{booktabs}
\begin{table}[h!]
\resizebox{\columnwidth}{!}{
\begin{tabular}{@{}ccccc|ccc@{}}
\toprule
\textbf{ClaimBuster}                & \multicolumn{2}{c}{\textbf{`groundtruth'}} & \multicolumn{2}{c}{\textbf{`crowdsourced'}} & \multicolumn{3}{c}{\textbf{NewsClaims}}                            \\ \midrule
\textit{assigned label}             & \textit{sent.}    & \textit{pct.}    & \textit{sent.}     & \textit{pct.}    & \textit{assigned label}     & \textit{sent.} & \textit{pct.} \\
Non-factual Sentence (NFS)          & 731               & 70.83                  & 15416                & 65.51              & Factually Verifiable Claims (FVC) & 655            & 12.70               \\
Unimportant Factual Sentence (UFS)  & 63                & 6.11                 & 2466                 & 10.48                  & non-Factually Verifiable Claims                      & 4502           & 87.29               \\
Check-worthy Factual Sentence (CFS) & 238               & 23.06                  & 5651                & 24.01                  & (non-FVC)                            &                &                     \\ \bottomrule
\end{tabular}%
}
\caption{Class distribution in datasets}
\label{distribution}
\end{table}

\section{Methodology}
 
We tested two kinds of models and various feature sets on the task. The first model type are statistical machine learning methods, which have shown good results for textual classification tasks, %TS: add reference??
particularly on small datasets. Here, we used lexical features, specific dictionaries, as well as several kinds of word embeddings and encodings of the claim metadata. The second model type are neural machine learning algorithms, which we tested with a BERT model and with ontology embeddings (as well as in combination).

\subsection{Vectorization in Statistical Machine Learning Methods}
The most common and successful algorithms building statistical machine learning approaches for text classification problems are Support Vector Machines and Logistic Regression. Vectorization is a crucial step to get  distinct features out of the text for these models to train on. We extracted naive binary term occurrence features (i.e. term frequency–inverse document frequency, TF-IDF) and enhanced them with Linguistic Inquiry and Word Count (LIWC) \citep{pennebaker2001linguistic} and linguistic features. Also, word embeddings and encoded metadata have been used as features for these statistical ML classifiers.

\paragraph{LIWC and other linguistic features}
As a baseline feature set, we used word token unigrams and bigrams along with TF-IDF.
In addition, we combined LIWC dictionary-based quantitative text analysis features with the n-grams. LIWC captures psychological processes and content-related aspects (e.g., positive or negative emotions, cognitive processes, social processes). We also included  several linguistic features such as word count, character count, punctuation count, text complexity and sentiment analysis scores.

\paragraph{Word2Vec for Feature Extraction}
We used a common word vector model, Word2Vec \citep{mikolov2017advances}, to take into account the semantic relatedness of tokens. There is a publicly available pre-trained Word2Vec model trained on the English CoNLL17 corpus with the skip-gram method\footnote{\url{http://vectors.nlpl.eu/repository/}}. In addition to this domain-free word embedding approach, we trained domain-specific word vectors on the training data via the gensim library. This takes  preprocessed data in the form of tokenized texts as input and produces a model as output. Text sequences from word embeddings have been created with vector aggregation. For this purpose, we used hash maps including words and their occurrences with defining certain length for each sequence. If the length of a sequence is shorter than the maximum length, we added zero padding, otherwise this sequence was trimmed from the end. 

\paragraph{Knowledge Graphs for Encoding of Metadata} 
Knowledge Graphs (KG) evolved as a technique for semantically structuring knowledge about real-world entities in a machine-readable format \citep{ehrlinger2016towards}. They are directed labeled graphs consisting of nodes, edges, and labels. Our motivation for employing KG structure is related to the challenging task, and to the domain dependence of fact verification. Some of the metadata already existing in the dataset constitutes nodes in the knowledge graph structure: Thus, the `speaker' is the name of the person who made the claim, `speaker title' is the speaker's occupation at the time of the debate, `speaker party' is the political affiliation of the speaker, as well as the number of words in the text and the sentiment of the text.  Additionally, we used the TagMe entity linking annotation API \citep{ferragina2011fast} to find the highest-scored entities in each claim. By encoding patterns between nodes, relations are created. After graph construction, we continued with graph embedding to obtain a low-dimensional representation of the feature space. The TransE model \citep{bordes2013translating} was used to embed the entities and relations by treating each relation as a transformation operation from the head entity to the tail entity (labeled with the relation). 

\subsection{Neural Network Based Approaches}
Neural network approaches based on sequence or graph modeling provide an opportunity for high performance in various NLP tasks. %Thus, we researched the neural network-based systems most suitable for the nature of our problem. 
Claim-related world knowledge was encoded using the advantage of the graph structure. The scores for our task were obtained by using the pretrained domain free language models Word2Vec and BERT. In order to achieve more successful results in a domain-specific task, we fused the BERT model with graph-embedded ontology entities.

\paragraph{Word2Vec Embedding}
As we mentioned in the previous section, publicly available pre-trained Word2Vec embeddings have been used. In this approach, the collection of embeddings was used to seed the embedding layer with the word embedding weights. These weights have been passed to a fully connected layer with 500 hidden units and the ReLU activation function. Then, the softmax activation function was performed for obtaining predictions. 

\paragraph{OWL2Vec* Ontology Embedding}
ClaimsKG is a structured database based on a knowledge graph model which includes 53,521 English claims published between the years 1996--2022 from 14 fact-checking websites  and was extended with DBpedia entities and schema.org and NIF vocabularies \citep{tchechmedjiev2019claimskg}. We scraped relevant data  from this database with the following features: claim bodies, normalized labels for claims, links for debunked news, the claim review body and its title, references cited in the claim review, the entities and their Wikipedia categories extracted from the claim body and from the review body, author of the claim review, the claim publication date. During data harvesting, non-English and corrupted data samples were not included. 

Different kinds of domain knowledge can be represented in an OWL ontology in structured form with an Abox part (classes, objects), a Tbox part (instances, facts) and its own syntax (W3C OWL, RDF, RDFS) \citep{antoniou2009web}. 
We created our own OWL ontology based on the scraped claim data and its metadata to obtain encoded claim-related tokens. %Thus, OWL ontology was implemented by following the data model and instances of ClaimsKG.
We implemented a new OWL ontology by following the data model and instances of ClaimsKG. 

For embedding the OWL ontologies, we used the OWL2Vec framework \citep{chen2021owl2vec}. It aims to represent each OWL named entity with a vector so that entity-related information is retained in vector space. \citet{chen2021owl2vec} emphasized that this approach has several advantages compared to other ontology embedding models such as combinations of the semantics of the graph structure, lexical information and logic constructors, different graph walking strategies, ontology entailment reasoning, and word embedding pre-training. Also, results of researchers’ case studies showed the impact of those features with achieving significantly better performance than state-of-the-art ontology embeddings (Onto2Vec, OPA2Vec), and knowledge graph embeddings (RDF2Vec, TransE). As in the previous model, we processed word vectors through a fully connected layer with 500 hidden units, and the Rectified Linear Unit (ReLU) activation function is applied. Lastly, predictions were generated using the softmax activation function.

\paragraph{BERT model}
We use the Transformer-based BERT model, which offers state-of-the-art solutions to numerous NLP problems. With its structure that bidirectionally processes the incoming text and combines the masked language and next-sentence prediction models, BERT is able to account for the context of a token. There is a publicly available  English language model provided by Google, that was trained on several English corpora such as the BookCorpus (a dataset consisting of 11,038 unpublished books) and English Wikipedia with 12 transformer blocks. We fine-tuned this uncased BERT model for the claim identification task. The recommended hyperparameters by \citet{devlin2018bert} were examined using the Adam optimizer and batch-size and common learning rate were chosen as 16 and 2e-5, respectively. 
%Sentence embeddings were constituted through obtaining the vectors corresponding to the [CLS] token from the final Transformer layer of this fine-tuned BERT model \\

\begin{figure}[!ht]
\centering
\includegraphics[scale=0.43]{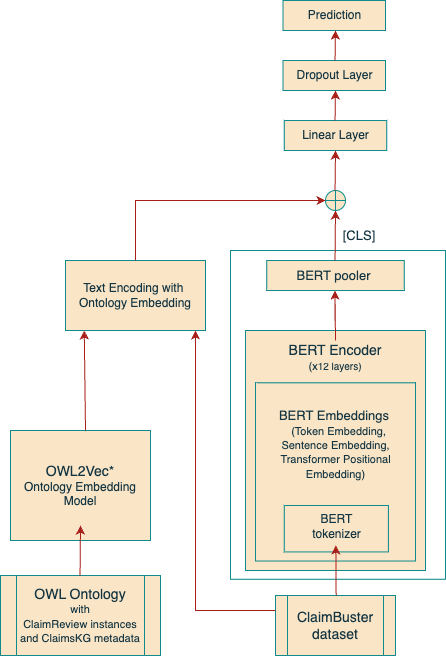}
\caption{Proposed model fusing textual and ontology features}
\label{figModel}
\end{figure}

\paragraph{Multimodal Fusion of BERT Model and Ontology Embeddings}
%TODO
The last hidden state of the CLS token serves as a good representation of input sequences. As shown in Figure~\ref{figModel}, we concatenated the CLS embedding with the corresponding ontology embeddings for each sequence vertically. These concatenated vectors go through the linear and dropout layer, and we use BCEWithLogitsLoss to produce scores for each label. 

\section{Experiments and Results}
\subsection{ClaimBuster dataset}
\paragraph{Small data experiments}
Our main experiment uses the small groundtruth data of ClaimBuster for both training and testing as specified above. 

For the sake of completeness and to examine the power of statistical ML models, especially in small and domain-specific datasets, we chose SVM and Logistic Regression classifiers with TF-IDF weighted word n-grams features as a baseline approach. According to our results shown in Table~\ref{statMLresult}, Logistic Regression obtains the highest scores in all metrics except the model that includes linguistic and LIWC features. The highest accuracy and weighted average F1-scores were obtained with the model leveraging knowledge graph embeddings. Also, instead of using pretrained embeddings, training Word2Vec embeddings on domain data reached the second-highest scores, despite the small size of the corpus. It can be seen that adding domain-specific features in the form of external knowledge or domain specific textual embeddings is useful for a context-based task such as claim identification.

Our experimental results of the neural network based approaches are presented in Table~\ref{tab:neuralMLresults}. \citep{jha2023towards} proposed G2CW framework that relies on Glove embeddings and gated recurrent units (GRU) that is the latest study on the ClaimBuster dataset as stated in Section \ref{related}. We replicated this model using the architecture and hyperparameters they specified, and applied it to the ClaimBuster groundtruth dataset. 

Our BERT model, which includes 12 transformer layers, achieved scores interestingly close to the fully connected, shallow neural network, GRU-based model, and statistical machine learning models for our task. The class-based results show that the BERT model did not produce any results for the UFS label, which has very few instances. The BERT model was biased for the class containing the largest number of samples in this small-sized and unbalanced dataset, and failed on the multiclass classification problem. On the other hand, the addition of ontology vectorization for corresponding text sequences to the BERT model (``BERT + Ontology Embeddings'') improved the accuracy by 0.06 points and macro-average F1-score by 0.19 points. Thus, our best model exhibits a macro-average F1-score of 0.74 and an accuracy of 0.92 on the sentence-level claim identification task. 

% \begin{figure}[!ht]
% \centering
% \small
% %\fbox{\parbox{6cm}{
% %This is a figure with a caption.}}
% % old picture \includegraphics[scale=0.5]{lrec2020W-image1.eps} 
% \includegraphics[scale=0.5]{histogram.jpg}
% \caption{Class based \emph{F1-scores} of neural network based models}
% \label{dist}
% \end{figure}

%\clearpage
\begin{table*}[htp]
\centering
\begin{tabular}{@{}llcccc|cccc@{}} 
\toprule
\multicolumn{2}{l}{\multirow{2}{*}{}}                                                                                        & \multicolumn{4}{c}{\textbf{SVM}}                                                                                         & \multicolumn{4}{c}{\textbf{Logistic Regression}}                                                                           \\ 
\cmidrule{3-6}
\cmidrule{7-10}
\multicolumn{2}{l}{\textit{ClaimBuster dataset (groundtruth)}}                                                                                                         & \textbf{P} & \textbf{R} & \textbf{F1} & \multicolumn{1}{c}{\textbf{Acc.}} & \textbf{P} & \textbf{R} & \textbf{F1} & \textbf{Acc.}  \\ 
\midrule
\multirow{2}{*}{word (1,2)-gram + tf-idf}                                                                      & macro avg    & 0.58                          & 0.55                       & 0.56                         & \multirow{2}{*}{0.81}         & 0.56                          & 0.49                       & 0.50                         & \multirow{2}{*}{0.83}         \\
 & wt. avg & 0.80                          & 0.81                       & 0.80                         &                               & 0.79                          & 0.83                       & 0.79                         &                               \\ 
\hline
\multirow{2}{*}{ling features + LIWC features}                                                                 & macro avg    & 0.61                          & 0.59                       & 0.59                         & \multirow{2}{*}{0.83}         & 0.53                          & 0.51                       & 0.52                         & \multirow{2}{*}{0.82}         \\
 & wt. avg & 0.82                          & 0.83                       & 0.82                         &                               & 0.78                          & 0.82                       & 0.80                         &                               \\ 
\hline
\multirow{2}{*}{pretrained Word2Vec features}                                                               & macro avg    & 0.65                          & 0.63                       & 0.64                         & \multirow{2}{*}{0.81}         & 0.71                          & 0.63                       & 0.66                         & \multirow{2}{*}{0.82}         \\
 & wt. avg & 0.81                          & 0.81                       & 0.81                         &                               & 0.81                          & 0.82                       & 0.82                         &                               \\ 
\hline
\multirow{2}{*}{\begin{tabular}[c]{@{}l@{}}training  Word2Vec features \\on domain data\end{tabular}}        & macro avg    & 0.69                          & 0.69                       & 0.69                         & \multirow{2}{*}{0.83}         & 0.74                          & 0.66                       & \bf 0.69                         & \multirow{2}{*}{0.84}         \\
 & wt. avg & 0.82                          & 0.83                       & 0.82                         &                               & 0.83                          & 0.84                       & 0.83                         &                               \\ 
\hline
\multirow{2}{*}{\begin{tabular}[c]{@{}l@{}}Knowledge Graph emb. of \\metadata + Word2Vec \end{tabular}} & macro avg    & 0.68                          & 0.69                       & 0.68                         & \multirow{2}{*}{0.84}         & 0.69                          & 0.67                       & 0.68                         & \multirow{2}{*}{0.85}         \\
 & wt. avg & 0.84                          & 0.84                       & 0.84                         &                               & 0.85                          & 0.85                       & \bf 0.85                         &                               \\
\bottomrule
\end{tabular}
\caption{Evaluation scores of the statistical machine learning based methods}
\label{statMLresult}
\end{table*}

\begin{table*}[htp]
\centering
\begin{tabular}{@{}llcccc@{}} 
\toprule
\multicolumn{2}{l}{\textit{ClaimBuster dataset (groundtruth)}}                                      & \textbf{Precision} & \textbf{Recall} & \textbf{F1} & \textbf{Accuracy}  \\ 
\midrule
\multirow{2}{*}{Word2Vec Embeddings + FCN}        & macro avg    & 0.69                                   & 0.62                                & 0.64                                  & \multirow{2}{*}{0.82}                  \\
                                            & weighted avg & 0.80                                   & 0.81                                & 0.81                                  &                                        \\ 
\hline
\multirow{2}{*}{Ontology Embeddings + FCN}        & macro avg    & 0.73                                   & 0.66                                & 0.69                                  & \multirow{2}{*}{0.85}                  \\
                                            & weighted avg & 0.84                                   & 0.84                                & 0.84                                  &                                        \\ 
\hline
\multirow{2}{*}{Glove Embeddings + GRU \citep{jha2023towards}}        & macro avg    & 0.62                                   & 0.63                                & 0.62                                  & \multirow{2}{*}{0.86}                  \\
                                            & weighted avg & 0.86                                   & 0.87                                & 0.86                                  &                                        \\ 
\hline
\multirow{2}{*}{BERT model}                 & macro avg    & 0.53                                   & 0.60                                & 0.55                                  & \multirow{2}{*}{0.86}                  \\
                                            & weighted avg & 0.94                                   & 0.86                                & 0.89                                  &                                        \\ 
\hline
\multirow{2}{*}{BERT + Ontology Embeddings} & macro avg    & 0.70                                   & 0.93                                & \bf 0.74                                  & \multirow{2}{*}{0.92}                  \\
                                            & weighted avg & 0.94                                   & 0.92                                & \bf 0.93                                  &                                        \\[.5ex]
\midrule\midrule 
\multicolumn{2}{l}{\textit{ClaimBuster dataset  (crowdsourced and groundtruth)}}                                      & \\%\textbf{Precision} & \textbf{Recall} & \textbf{F1} & \textbf{Accuracy}  \\ 
\midrule
\multirow{2}{*}{BERT model}        & macro avg    & 0.91                                   & 0.95                                & 0.92                                  & \multirow{2}{*}{0.97}                  \\
                                            & weighted avg & 0.97                                   & 0.97                                & 0.97                                  &                                        \\ 
\hline
\multirow{2}{*}{BERT + Ontology Embeddings}        & macro avg    & 0.91                                   & 0.94                                & 0.92                                  & \multirow{2}{*}{0.97}                  \\
                                            & weighted avg & 0.97                                   & 0.97                                & 0.97                                  &                                        \\ 
\midrule\midrule 
\multicolumn{2}{l}{\textit{NewsClaims dataset}}                                      & \\%\textbf{Precision} & \textbf{Recall} & \textbf{F1} & \textbf{Accuracy}  \\ 
\midrule
\multirow{2}{*}{BERT model}        & macro avg    & 0.44                                   & 0.5                                & 0.47                                  & \multirow{2}{*}{0.87}                  \\
                                            & weighted avg & 0.76                                   & 0.87                                & 0.81                                  &                                        \\ 
\hline
\multirow{2}{*}{BERT + Ontology Embeddings}        & macro avg    & 0.83                                   & 0.59                                & \bf 0.62                                  & \multirow{2}{*}{0.89}                  \\
                                            & weighted avg & 0.88                                   & 0.89                                & \bf 0.86                                  &                                        \\ 
\bottomrule
\end{tabular}
\caption{Evaluation scores of neural network based approaches}
\label{tab:neuralMLresults}
\end{table*}

\paragraph{Large data experiments}
The large crowdsourced part of the ClaimBuster dataset as the training data and the groundtruth part as the test data were used for comparison. We experimented with our best two models, the fine-tuned BERT model and the ontology enhanced BERT model. We performed the experiments with the hyperparameters that we used in the previous BERT-based models (i.e. Adam optimizer, learning rate 2e-5, batch-size 16). These two models obtained almost identical results and reached an accuracy of 0.97, and macro-average F1-score of 0.97, also shown in Table \ref{tab:neuralMLresults}. It can be seen that a large annotated in-domain dataset for training can compensate for the disadvantages of the plain BERT model, or the reverse, that external domain knowledge in the shape of ontologies can be leveraged to compensate for a small training data set, when combined with the textual information. 

\subsection{NewsClaims dataset}
To examine the effect of integrating ontology information in the claim identification task for a small-sized dataset, we applied our proposed ontology leveraged BERT model and the fine-tuned BERT model to the NewsClaims data by following the same hyperparameter set. The best results for this dataset, which requires in-depth understanding, were obtained by the claim ontology enhanced BERT model with an accuracy score of 0.89 and a macro average score of 0.62. This result is compatible with the ClaimBuster groundtruth dataset with similar features.

\section{Discussion and Conclusion}
In this study, we approached the claim detection problem by incorporating domain knowledge along with linguistic features in different machine learning models. The effect of different feature sets on the small-size datasets was examined. While graph-structured features can capture background knowledge, text can capture context. We show that both modalities of information must be combined as they can enrich each other. Ontology models have been shown to be appropriate for the detection of factual check-worthiness, an implicit and context-dependent problem. These models benefit from the ability to link together pieces of information that enable automated reasoning to extract information that has not been explicitly mentioned before. 

We encoded fact-check related data from an internet knowledge base and adapted it to our problem by converting it into an ontology and embedding the resulting ontology. Our proposed ontology enhanced claim detection model achieved the highest scores in all metrics for the groundtruth part of the ClaimBuster and NewsClaims datasets. This model also proved that adding ontology features can avoid the bias encountered in neural network based models, for example the pure BERT model's bias towards larger classes. %The ontology enhanced model showed a good performance on our small dataset.
Besides this task, the embedded fake news-related ontology data structure can also be effective in various fact-checking problems, with its structured data shape, including the relationship between entity and news texts, and the Wikipedia data it contains.

Our conducted experiments also show that domain-specific features like a knowledge graph embedding of metadata,  word embeddings trained on in-domain data, and taking advantage of a fact check-related OWL ontology to encode input sequences can be more effective than domain free features like pretrained language models Word2Vec, Glove and BERT. 

We further examined the effect of the data size and the addition of ontology features to a large-size dataset for the BERT model, by retraining it on the crowdsourced part of ClaimBuster and testing on the groundtruth part. We concluded that the domain-specific feature addition is not a very effective solution to reach higher scores for the large-size data. But those features can have a significant impact in the absence of an expert annotated reliable dataset of a reasonable size, such as in the checkworthiness task for languages other than English, where data is very sparse. In addition, domain-specific features and graph-structured data that may achieve similar scores to neural network models can play an important role in the development of transparent and human-understandable machine-learning systems.

\bibliography{anthology,custom}
\appendix

\section{Error Analysis}
In order to investigate the impact of the ontology information in detail, we evaluated the instances of the  ClaimBuster groundtruth dataset  where the two main models (ontology enhanced versus pure BERT) disagreed in the labels; statistics for these cases are presented in Table ~\ref{errorstat}. It can be observed that when the ontology enhanced BERT model correctly predicted the samples belonging to the UFS class, the BERT model gave wrong results for 4 samples belonging to this class. In the CFS class, there were 14 samples in which the BERT model predicted incorrectly and the ontology enhanced BERT model made a correct prediction. Contrary to this, there was no example where the BERT model gave the correct result while the ontology enhanced BERT model was incorrect in the UFS and CFS classes. The class in which the test results of the models are closest to each other is NFS with the highest number of samples.

\begin{table}[h!]
\small
\centering
\begin{tabular}{@{}crrr@{}}
\toprule
\textbf{(\# of samples in the error set)} &\textbf{UFS}  & \textbf{CFS} & \textbf{NFS}\\
\midrule
\begin{tabular}[c]{@{}c@{}}BERT + Ontology Emb. \textbf{correct},\\ BERT model \textbf{incorrect}  \end{tabular} &4&14& 5\\
\begin{tabular}[c]{@{}c@{}}BERT + Ontology Emb. \textbf{incorrect},\\ BERT model \textbf{correct}  \end{tabular} & 0 & 0 & 4\\
\bottomrule
\end{tabular}
\caption{Error distribution for the ClaimBuster groundtruth dataset, models with and without ontology}
\label{errorstat}
\end{table}

We chose 6 sample sentences from the test set errors for further visual analysis. %This sample set contains two examples from each class UFS, CFS and NFS. 
To take a closer look at the failure of the plain BERT model to assign the UFS class, we compare two instances from UFS (which were correctly labelled by BERT+Ontology, but mislabelled by plain BERT) with two typical samples from each of the CFS and NFS classes, which were correctly assigned by both models. Tables ~\ref{attenBERT}--\ref{attenBERTonto} show these sentences with a visualization of the degree of attention given to each token by the plain BERT model (Table \ref{attenBERT}) and the BERT+Ontology model (Table \ref{attenBERTonto}). Sentence embeddings of these samples were formed by averaging the weights from the last 4 layers for the two models. The shades of colors represent the degree of attention and darker words have higher weight scores. While quantity words (i.e., `three days', `six predecessors') have higher weights in the BERT model, it can be seen that central entities (i.e., `president', `Senator Obama', `Oval Office', `women's rights') are more important for us to decide whether a claim will be checkworthy or not in the ontology enhanced BERT model. In the sentences that the BERT model predicts incorrectly, the BERT model assigns a higher weight value to the first-person pronouns. It can be considered that the use of such features is associated with a more informal and spontaneous style of discourse which may be important for the determination of the claim's checkworthiness.

%By following the same approach, we examined the results of our proposed BERT-based models on the NewsClaims dataset. The colorized weight attention scores for these two models are presented in Tables \ref{attenBERTnewsclaim}--\ref{attenBERTontonewsclaim}. In the fine-tuned BERT model, the weights of the samples did not differ significantly between tokens. The ontology enhanced BERT model gives more importance to entities related to the main topic of the dataset (i.e., `coronavirus').  

%In addition, Figure \ref{sec:appendix} (provided in the Appendix) shows the sentence embeddings of these six samples in two-dimensional space, obtained with principal component analysis (PCA). Each class sentence was colored differently. It can be observed that the BERT model has difficulty in distinguishing the smaller UFS and CFS classes, whose sentence representations are close together.
%In the ontology enhanced BERT model, there is a clearer spatial distinction between sentence embeddings for instances from all three classes. 

\begin{table}[h!]
\resizebox{\columnwidth}{!}{
     \begin{tabular}{ c  p{2cm} p{2cm}  }
     \toprule
      sample sentences from test set of ClaimBuster (groundtruth) & true class & BERT model prediction \\ 
     \midrule
     \raisebox{-\totalheight}{\includegraphics[width=0.7\textwidth]{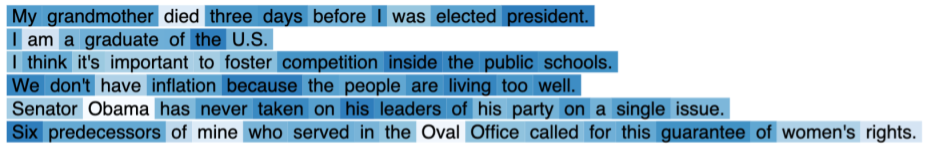}} & 
      \begin{itemize}
      \vspace{-3mm}
      {\tiny \item UFS}
      {\tiny \item UFS}
      {\tiny \item NFS}
      {\tiny \item NFS}
      {\tiny \item CFS}
      \vspace{-1.5mm}
      {\tiny \item CFS}
      \vspace{-1.5mm}
      \end{itemize} & 
      \begin{itemize}
      \vspace{-3mm}
      {\tiny \item \textbf{CFS}}
      {\tiny \item \textbf{CFS}}
      {\tiny \item NFS}
      {\tiny \item NFS}
      {\tiny \item CFS}
      \vspace{-1.5mm}
      {\tiny \item CFS}
      \vspace{-1.5mm}
      \end{itemize} \\
      \bottomrule
      \end{tabular}}
\caption{Highlighted words by considering weights from the BERT model}
\label{attenBERT}
\end{table}

\begin{table}[ht!]
\resizebox{\columnwidth}{!}{
     \begin{tabular}{ c  p{2cm} p{2.5cm}   }
     \toprule
      sample sentences from test set of ClaimBuster (groundtruth)& true class & BERT+Ontology prediction  \\ 
    \midrule
     \raisebox{-\totalheight}{\includegraphics[width=0.7\textwidth]{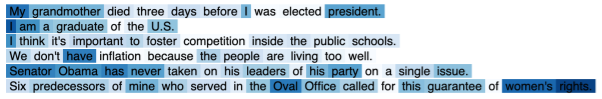}}
      & 
      \begin{itemize}
      \vspace{-3mm}
      {\tiny \item UFS}
      {\tiny \item UFS}
      {\tiny \item NFS}
      {\tiny \item NFS}
      {\tiny \item CFS}
      \vspace{-1.5mm}
      {\tiny \item CFS}
      \vspace{-1.5mm}
      \end{itemize}
      & 
      \begin{itemize}
      \vspace{-3mm}
       {\tiny \item  UFS}
      {\tiny \item UFS}
      {\tiny \item NFS}
      {\tiny \item NFS}
      {\tiny \item CFS}
      \vspace{-1.5mm}
      {\tiny \item CFS}
      \vspace{-1.5mm}
      \end{itemize}
      \\ \bottomrule
      \end{tabular}}
\caption{Highlighted words by considering weights from the ontology enhanced BERT model}
\label{attenBERTonto}
\end{table}

\end{document}